\documentclass[letterpaper, 10 pt, conference]{ieeeconf}  

\IEEEoverridecommandlockouts                              

\usepackage{graphics} 
\usepackage{times} 
\usepackage{amsmath} 
\usepackage{amsfonts}
\usepackage{multicol}
\usepackage{graphicx}
\usepackage{caption}
\usepackage{subcaption}
\usepackage{gensymb}

\usepackage{booktabs}
\usepackage[bookmarks=true]{hyperref}
\newcommand{\norm}[1]{\lVert#1\rVert_2}

\usepackage{xcolor}
\usepackage{esvect}

\DeclareMathOperator*{\argmin}{arg\,min}

\usepackage{color, soul}
\renewcommand\hl[1]{#1} 

\title{\LARGE \bf
Cloth Region Segmentation for Robust Grasp Selection
}

\author{Jianing Qian$^{*}$, Thomas Weng$^{*}$, Luxin Zhang, Brian Okorn, David Held
\thanks{Jianing Qian, Thomas Weng, Luxin Zhang, Brian Okorn, and David Held are with the Robotics Institute, Carnegie Mellon University, Pittsburgh, PA.
        {\tt\footnotesize \{jianingq, tweng, luxinz, bokorn, dheld\}@andrew.cmu.edu}}%
\thanks{*These authors contributed equally and are listed in alphabetical order.}%
}

\begin{document}

\maketitle
\thispagestyle{empty}
\pagestyle{empty}


\begin{abstract}
Cloth detection and manipulation is a common task in domestic and industrial settings, yet such tasks remain a challenge for robots due to cloth deformability. Furthermore, in many cloth-related tasks like laundry folding and bed making, it is crucial to manipulate specific regions like edges and corners, as opposed to folds. In this work, we focus on the problem of segmenting and grasping these key regions.
Our approach trains a network to segment the edges and corners of a cloth from a depth image, distinguishing such regions from wrinkles or folds. We also provide a novel algorithm for estimating the grasp location, direction, and directional uncertainty from the segmentation. We demonstrate our method on a real robot system and show that it outperforms baseline methods on grasping success. Video and other supplementary materials are available at: \mbox{\href{https://sites.google.com/view/cloth-segmentation}{https://sites.google.com/view/cloth-segmentation}}.
     
\end{abstract}

\section{Introduction}
\label{sec:intro}

Manipulating and interacting with cloth is a key part of daily life, yet cloth manipulation by robots remains a challenging problem. 
Cloth is difficult to perceive and manipulate because its deformable nature breaks the rigid-body assumptions of many algorithms. For example, most pose estimation algorithms assume that objects can only transform in 6 degrees of freedom (translation and rotation). However, cloth can deform at any location and thus has nearly an infinite number of degrees of freedom.

In cloth-based tasks like laundry folding and textile manufacturing, it is important to detect and grasp specific regions of cloth, e.g. corners and edges, for downstream manipulation like folding or smoothing. These edges and corners are distinct from wrinkles and folds, which are less useful for downstream tasks.

In order to grasp the cloth along an edge or corner, we must not only detect the cloth edges and corners but also estimate the appropriate grasping direction.  Given a grasp position, the grasp direction specifies the approach vector the gripper follows towards this point. Although estimating the grasping direction would be relatively simple if the cloth were lying flat on the table, it is much more challenging in crumpled configurations.
Much work has been done for perception and manipulation of cloth in both randomized and predefined cloth configurations, yet cloth-related tasks like laundry folding and assisted dressing remain challenging due to the inherent complexity of cloth dynamics.

\begin{figure}[t]
    \centering
    \includegraphics[width=\columnwidth]{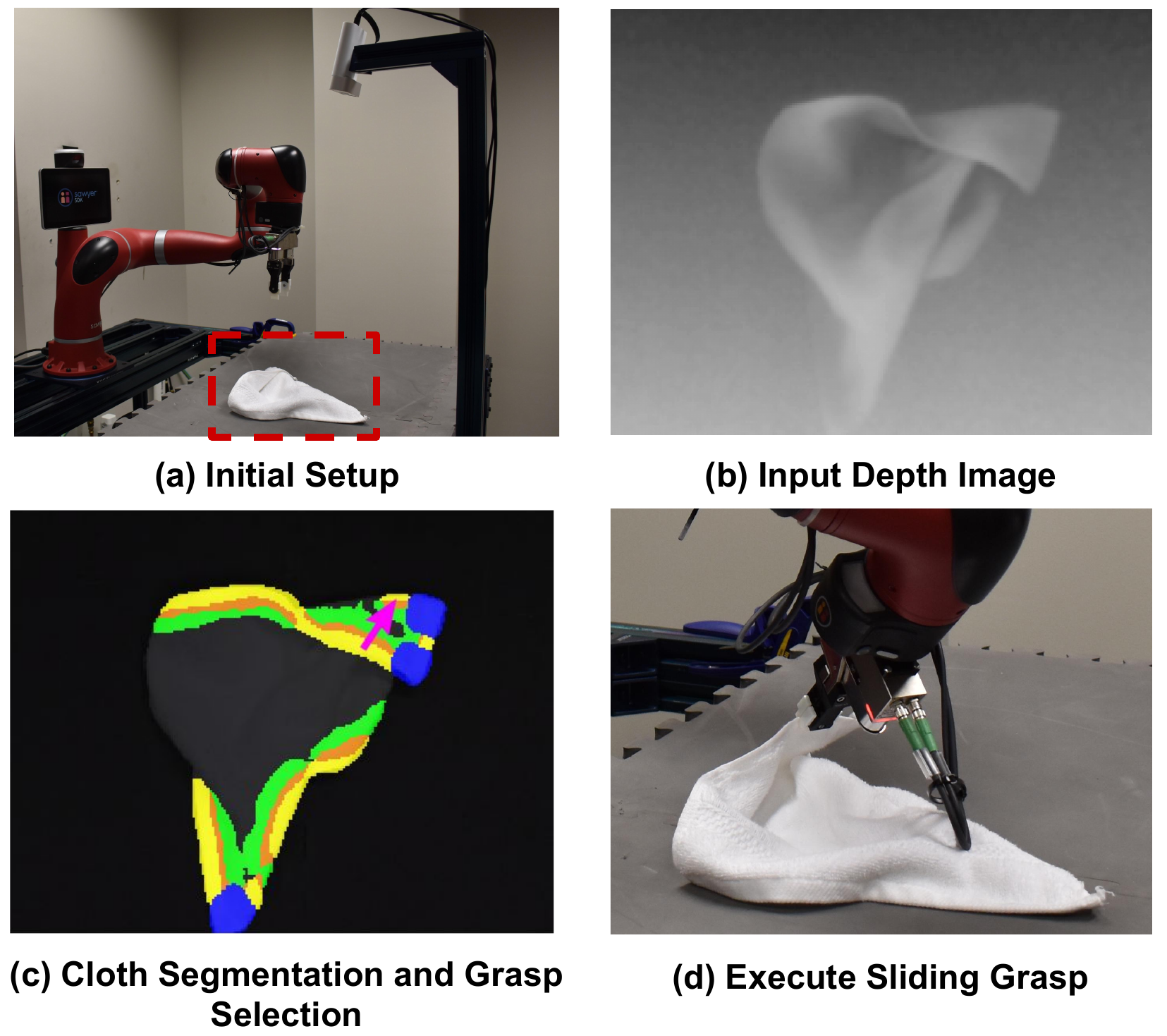}
    \caption{Grasping using cloth region segmentation: Robot with depth sensor (a) captures depth image of test cloth (b). Depth image is segmented into outer edges (\colorbox{black!45}{\textcolor{yellow}{yellow}}), inner edges (\textcolor{green}{green}) and corners (\textcolor{blue}{blue}) using our cloth region segmentation network (c). Ambiguous regions are colored in \textcolor{orange}{orange}. Our method selects a grasp location and direction, shown as a \textcolor{magenta}{magenta} arrow. The robot executes a sliding grasp and successfully grips the cloth by its edge.} 
    \label{fig:overview}
\end{figure}

In this paper, we present an approach for segmenting these key regions of cloth, even in highly crumpled configurations. 
To achieve this, we train a neural network to predict cloth edges and corners from a depth image. We also train the network to predict the inner edges, the region interior to the cloth's true edges, for grasp direction estimation.
The network is trained on a dataset of RGB-D images extracted from 8 minutes of video of a human manipulating the cloth. 
The ground-truth for the network is provided by color-labeling the cloth (see Fig.~\ref{fig:overview}), forgoing the need for expensive human annotations. 

The segmentation output of our network allows us to quickly and robustly estimate the appropriate position and grasp direction from a crumpled cloth. It also allows us to estimate the grasp directional uncertainty for every edge/corner pixel. This estimation is important for grasping the cloth, as mis-estimating the grasp direction and approaching at an angle not orthogonal to the cloth edge is more likely to fail.
Using a dense estimate of grasp directional uncertainty, we can choose the grasp point most likely to succeed.

We implement our method on a real robot system and evaluate its performance on grasp success metrics against a number of baselines; this evaluation demonstrates the strength of our system in estimating cloth edge and corner positions, grasp direction, and grasp uncertainty.

Our contributions include:
\begin{itemize}
    \item A method to segment regions of cloth critical for downstream manipulation tasks.
    \item An algorithm to determine a robust grasp configuration accounting for uncertainty about the cloth direction.
    \item An evaluation of our method against baselines on a real robot system for grasping edges and corners of cloth in crumpled configurations.
\end{itemize}

\section{Related Work}
\label{sec:related}



\subsection{Cloth Perception}

Robotic cloth manipulation is a well-studied domain with a variety of unsolved tasks, including laundry folding~\cite{maitin2010cloth,bimanualfolding}, laundry unfolding or smoothing~\cite{unfolding,fgbgunfolding,geounfolding,interactiveunfolding,forestunfolding,unfoldinggripping}, bed making~\cite{Laskey2017LearningRB,seita_bedmake_2019}, and grasping~\cite{demura2018picking,singlearmtowel,Wu2019LearningTM}.

Many of these approaches use traditional computer vision algorithms to detect cloth regions for various downstream tasks:~\cite{interactiveunfolding} chooses candidate grasp points by using Harris corner detection and discontinuity checks on the depth image for peak ridges and peak corners. \cite{unfolding} uses a pre-task manipulation, lifting the towel into the air and shaking it to remove wrinkles before returning it to the table. Canny edge detection is then used to compute contours for interior and exterior corner classification. \cite{maitin2010cloth} performs background subtraction and uses stereo images to select a centered point in a pile of towels. They grasp the towel from a central point and and rotate it to obtain a sequence of images. Towel corners are fit to these images using RANSAC. These perception algorithms usually require significant pre-manipulations to get a more structured configuration of the cloth, thus they are more time consuming than many learning-based methods. Furthermore, without these pre-manipulations, these methods are likely to fail under difficult initial configurations, such as highly crumpled cloth. We will show in Sec.~\ref{sec:expresults} that our method is much more robust to these crumpled cloth configurations compared to these traditional methods.

Another group of methods apply learning-based algorithms for image feature extraction. \cite{demura2018picking} uses the YOLO detection network to detect the thickest folded edge and grasp a folded towel from a stack. \cite{unfoldinggripping} uses an autoencoder network to predict the real edges of towels. This is similar to our approach; however, their method trains a network to output latent features and performs nearest-neighbor classification on input features to predict good grasp points, whereas our network \textit{directly} outputs segmentation masks of grasp regions and also determines good \textit{grasp directions}.
Their method also operates on RGB images and requires a human-annotated dataset of corners, whereas our method takes depth images as input to be invariant to changes in visual texture, and does not require human labeling. 

The most similar method to ours is~\cite{seita_bedmake_2019} which learns to identify a corner of a bed sheet by painting the corner red. \hl{Our method expands upon this work by estimating a \textit{dense segmentation} of multiple real edges, inner edges, and corners, as opposed to regressing to a \textit{single} corner position}. Furthermore, our method outputs dense grasp direction proposals as well as their corresponding uncertainty estimates. As we will show in Sec.~\ref{sec:expresults}, the grasp direction proposals and uncertainty estimates are crucial for our performance on our grasping evaluation. Specifically, these two outputs enable us to handle challenging crumpled cloth configurations.

\subsection{Cloth Grasping}

Although the focus of our work is on perception rather than grasping, we review prior work on cloth grasping strategies. A simple top-down or angled grasp is commonly used once a grasp point has been selected~\cite{interactiveunfolding,seita_bedmake_2019}.  A top-down grasp followed by a 6DOF grasping on detected corners of the the hanging cloth has also been studied~\cite{maitin2010cloth}. 

 Other prior works learn a policy for grasping.~\cite{singlearmtowel} learns parameters for motion and grasp primitives to grasp a folded towel. 
 ~\cite{demura2018picking} uses Q-learning to train a policy for grasping a folded towel from a stack.~\cite{Wu2019LearningTM} uses Soft-Actor-Critic to train a policy for rope and cloth manipulation. 

In our work, we identify the real corners and edges of the cloth and select a robust grasping point.
Then we execute a hand-designed sliding grasp policy on the selected grasping point in order to pick up the cloth by a single edge or corner.

\section{Approach}
\label{sec:approach}

\subsection{Problem Statement}
In cloth manipulation tasks such as laundry folding, it is important that the robot be able to identify and grasp key regions of the cloth. These regions typically include the ``real edges" or corners of a cloth. By ``real edges," we mean the edges of the cloth in the unfolded configuration, as opposed to any folds or creases that may appear as edges in a particular configuration.
If the robot grasps a cloth fold or crease and attempts to use such a grasp to neatly fold the cloth, the result likely will not end up as expected.  Thus, failing to grasp the cloth along the real edges could lead to failures for many downstream tasks. 

As we will show, traditional computer vision algorithms fail to distinguish the difference between a real cloth edges and apparent edges created by creases or folds. In addition, the robot must also determine the appropriate grasping direction along the cloth edge, which is non-trivial if the cloth is in a crumpled configuration; we will show that simple heuristics frequently fail at this task. In this section, we provide a method that identifies edges and corners of a cloth, predicts grasp directions, and estimates the uncertainty of these directions. These predictions will then be used to quickly and reliably grasp the cloth along its edges and corners, even from crumpled configurations.

\subsection{Method Overview}
\label{sec:methodoverview}

\begin{figure*}[h]
    \centering
    \includegraphics[width=0.99\textwidth]{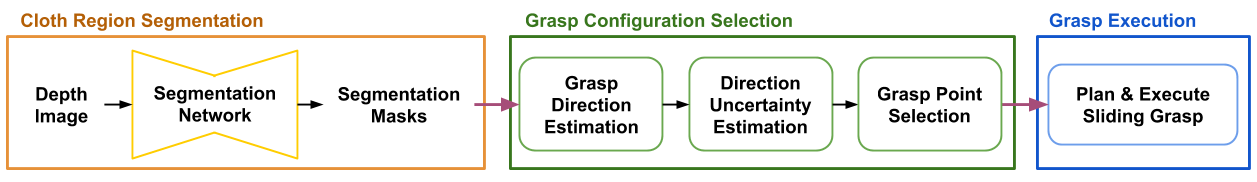}
    \caption{Pipeline for our method.
    Cloth region segmentation takes a depth image and outputs segmentation masks for cloth edges and corners. Grasp selection uses the masks to compute a grasp point and direction in the camera frame. Grasp execution transforms the grasp configuration into the robot frame and executes the grasp.}
    \label{fig:pipeline}
\end{figure*}

Fig.~\ref{fig:pipeline} provides the overall pipeline of our method. First, our segmentation network takes in a depth image and predicts the outer edges, inner edges and corners. Based on the segmentation, we estimate the grasp direction by computing a correspondence between outer edge and inner edge points. Next, we compute the grasp direction estimation and select a grasp point based on our uncertainty estimate $\mathbf{U(p)}$ for an outer edge point $\mathbf{p}$. Finally, we estimate the 6D robot pre-grasp pose based on the grasp point selected and execute our sliding grasp policy. These components are explained in greater detail in the following sections.

\subsection{Cloth Region Segmentation}
\label{sec:clothseg}

We frame the problem of identifying important regions cloth as semantic segmentation.  We train a neural network which receives as input a depth image of the scene containing the cloth.  The network predicts semantic labels for each pixel, giving the probability that the pixel contains a cloth outer edge, inner edge, corner, or none of these.  We can then threshold this probability to obtain a semantic segmentation mask for the cloth edge and corner locations. Fig.~\ref{fig:overview}c shows an example output of our network. 

\begin{figure}[ht]
    \centering
    \includegraphics[width=0.99\columnwidth]{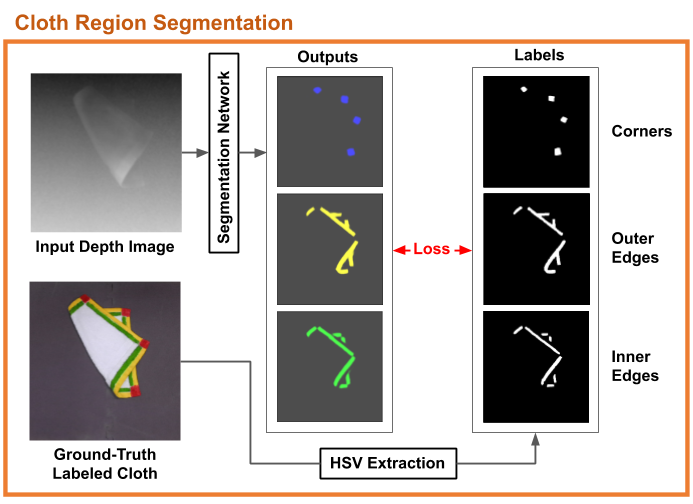}
    \caption{Training the segmentation network. The network receives a depth image as input. A paired RGB image supervises the network through the color labels of the cloth. Different colors are used to label the corners, outer edges, and inner edges. The ground-truth color for corner labels was changed from red to blue in the outputs to be color-blind friendly.}
    \label{fig:training}
\end{figure}

To train such a network, we need ground-truth labels for the cloth edges and corners.  Unfortunately, these are difficult to obtain in images with crumpled cloth, as this would require a large amount of human annotation effort. Instead, we adopt an approach similar to that of~\cite{seita_bedmake_2019}, in which they mark a single corner of a cloth with a red marker, and train a network to regress to the single corner location. In our case, we mark all edges and corners with different colors of paint and set up the problem as semantic segmentation, to estimate the position of all cloth edges and corners in the image (other differences from~\cite{seita_bedmake_2019} are explained in Sec.~\ref{sec:related}). 

As we will show, these labels will allow our network to differentiate between real edges or corners of the cloth from cloth folds, which may appear similar to edges in an image. Fig.~\ref{fig:training} is a visualization of our training method.

\hl{To train the segmentation network parameters $\theta$ using these labels, we define the loss $\mathcal{L}$ to be the mean of the pixel-wise binary cross-entropy loss $\ell_k$ for each class $k \in K$:} 
\begin{subequations}
\begin{equation}
\mathcal{L}(\theta) =  \frac{1}{K}\sum_k^K \ell_k
\end{equation}
\begin{equation}
\ell_k = -\sum_{i \in I} w_k (y_i \log\hat{y}_i) + (1 - y_i) \log (1 - \hat{y}_i)
\end{equation}
\end{subequations}
\hl{where $i$ is a pixel in the input depth image $I$, $w_k$ is a per-class weight to handle the imbalanced distribution between positive and negative labels, $y_i$ is the binary pixel label, and $\hat{y}_i$ is the network prediction for pixel $i$.}

\subsection{Grasp Configuration Selection}
\label{sec:graspconfig}

\begin{figure}[ht]
    \centering
    \vspace{0.2em}
    \includegraphics[ width=0.99\columnwidth]{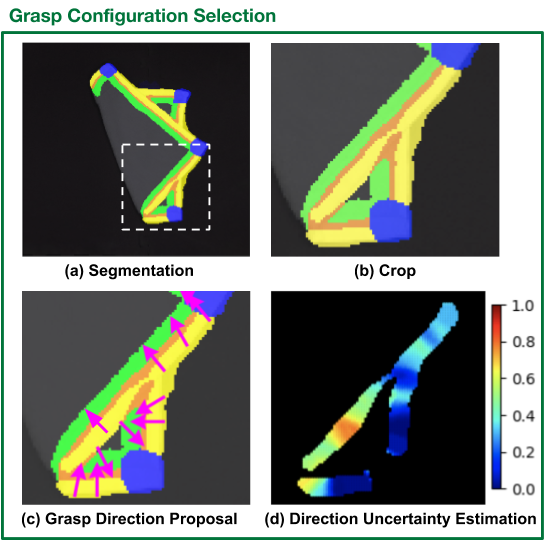}
    \caption{Illustration of grasp configuration selection. Corners are labeled in \textcolor{blue}{blue}, outer edges in \colorbox{black!45}{\textcolor{yellow}{yellow}}, inner edges in \textcolor{green}{green}. Overlapping outer edge and inner edge segmentations are in \textcolor{orange}{orange}; After obtaining the cloth region segmentation, (b) shows the cropped section in (a); (c) shows a subsample of  grasp direction proposals for each outer edge points; (d) shows the grasp directional uncertainty for each outer edge points.}
    \label{fig:correspondence}
\end{figure}

\subsubsection{Grasp Direction Estimation}
\label{sec:graspdir}

Once the edges and corners are estimated, the next step is to determine the appropriate grasp direction.  To achieve this, we augment the above pipeline by also predicting the cloth ``inner edges." We define the cloth outer edge as the region within 1.5 cm of the cloth edge, the cloth corners as the region within 3$\times$3 cm of the corner, and the inner edge as a 1.5 cm region interior to the cloth outer edge. The inner edge labels are shown in green in Fig~\ref{fig:training}. As before, we obtain cloth inner edge ground-truth labels using another color to paint the inner edge of a cloth, and we train a neural network to predict the cloth inner edge from a depth image.

Given the predicted segmentation for these cloth regions, we now select a grasp point and direction. 
We want to select the direction that allows our sliding grasp policy to most easily grasp the cloth. 
A sliding grasp that starts  with the gripper oriented towards a cloth edge as in Fig.~\ref{fig:expsetup} will intercept the edge upon translation.
However, a grasp oriented parallel to the edge or approaching from the reverse direction will not intercept the edge and will fail to grasp. 
Grasp direction is similarly important for corners, as sliding grasps that approach the corner head-on or aligned with the edge of the cloth are more likely to succeed than other orientations. 

The following is our procedure for computing the appropriate grasp direction. We first threshold the output of the network described in Sec.~\ref{sec:clothseg} to obtain a set of points estimated to belong to the outer edge $\mathbf{E_O}$ and a set of points that belong to the inner edge $\mathbf{E_I}$.  Then, for each outer edge point $\mathbf{p}=[p_x,p_y] \in \mathbf{E_O}$, we find the closest inner edge point $\mathbf{q}^*=[q_x,q_y]$. More formally, we define $\mathbf{q}^*$ to be 
\begin{alignat}{2}
&\mathbf{q}^* = \argmin_{\mathbf{q} \in \mathbf{E_I}}  \, \norm{\mathbf{p}-\mathbf{q}}  \label{eq:optProb}
\end{alignat}
With the correspondence between $\mathbf{p}$ and $\mathbf{q}^*$, we further define the grasp direction at point $\mathbf{p}$ to be the direction along the vector from $\mathbf{p}$ to $\mathbf{q}^*$.
Fig.~\ref{fig:correspondence}c shows a subset of those grasp directions. The vector from $\mathbf{p}$ to $\mathbf{q}^*$ often defines an appropriate grasp direction at point $\mathbf{p}$. This direction can be used by the robot to grasp the cloth.

\subsubsection{Directional Uncertainty Estimation}
\label{sec:diruncertainty}

Fig.~\ref{fig:correspondence}c also shows a few cases where, due to the complex folds of the cloth, the vector from $\mathbf{p}$ to $\mathbf{q}^*$ does not indicate an appropriate grasp direction.  Thus, for robust grasping, we also compute a measure of the uncertainty in this grasp direction.

We define the uncertainty of the grasp direction for a single point $\mathbf{p}$ to be the variance of the grasp directions predicted by its neighbours. To compute this variance, let $\mathbf{N_k(p)}$ be the set of $k$ closest pixel points in $\mathbf{E_O}$ of $\mathbf{p}$ in Euclidean distance; let $\alpha$ be the angle between $\vv{\mathbf{pq}^*}$ and a unit vector along the horizontal x axis. Formally we can define the cosine and sine of the grasp direction at $\mathbf{p}$ as
\begin{align}
f_{cos}(\mathbf{p}) &= cos(\alpha) = \frac{q_x-p_x}{\norm{\mathbf{q}^*-\mathbf{p}}}\\
f_{sin}(\mathbf{p}) &= sin(\alpha) = \frac{q_y-p_y}{\norm{\mathbf{q}^*-\mathbf{p}}}
\end{align}
We can then define observation vectors $\mathbf{x_0(p)}$ and $\mathbf{x_1(p)}$ to contain the cosine and sine of the grasp direction of all points in $\mathbf{N_k(p)}$:
\begin{align}
\mathbf{x_0(p)} &= \Big\{ f_{cos}(n) \mid n \in \mathbf{N_k(p)}\Big\}\\
\mathbf{x_1(p)} &= \Big\{ f_{sin}(n) \mid n \in \mathbf{N_k(p)}\Big\}
\end{align}
Next, we define the sample covariance matrix $\mathbf{K(p)}$ in the usual manner from the observations $\mathbf{x_0(p)}$ and $\mathbf{x_1(p)}$
\begin{align}
  \mathbf{K}_{ij}(p) =\frac{1}{N-1}\sum_{k=1}^{N}\left(  x_{ik}(p)-\bar{x}_i(p) \right)  \left( x_{jk}(p)-\bar{x}_j(p) \right)  
\end{align}
where $x_{ij}(p)$ is the $j$th element of $\mathbf{x_{i}(p)}$, and $\bar{x}_i(p)$ is the mean of $\mathbf{x_{i}(p)}$.

Finally, we define the uncertainty of our grasp direction prediction to be the sum of the variances of the individual dimensions, or the trace of $\mathbf{K}$:
$Tr(\mathbf{K(p)}) = Var(\mathbf{x_0(p)}) + Var(\mathbf{x_1(p)})$, where $Var(\mathbf{x_i(p)})$ is the variance of $\mathbf{x_i(p)}$. 
Since the trace of a matrix is equal to the sum of its eigenvalues, this means that $Tr(\mathbf{K})$ measures the summation of the uncertainty in the principal directions for the covariance matrix $\mathbf{K}$.
The trace therefore captures the uncertainty of the grasp direction while being invariant to axis transformations. Fig.~\ref{fig:correspondence}d shows an example of our uncertainty estimate.

\subsubsection{Grasp Point Selection}
\label{sec:grasppoint}
Finally, we describe our method for grasp point selection, which considers the outer edge predictions of Sec.~\ref{sec:clothseg} and the directional uncertainty estimates of Sec.~\ref{sec:diruncertainty}. For each outer edge point $\mathbf{p} \in \mathbf{E_O}$, we compute an uncertainty estimate  $\mathbf{U(p)} = Tr(\mathbf{K(p)})$ as described above.  Finally, for grasp point selection, we pick the outer edge point 
$\mathbf{p}$ that has the lowest uncertainty:
\begin{equation}
\mathbf{p} = \argmin_{\mathbf{p} \in \mathbf{E_O}} \, \mathbf{U(p)} \label{eq:optProb}
\end{equation}

\begin{figure}[h]
    \centering
    \begin{subfigure}[t]{0.32\columnwidth}
        \includegraphics[width=0.99\linewidth]{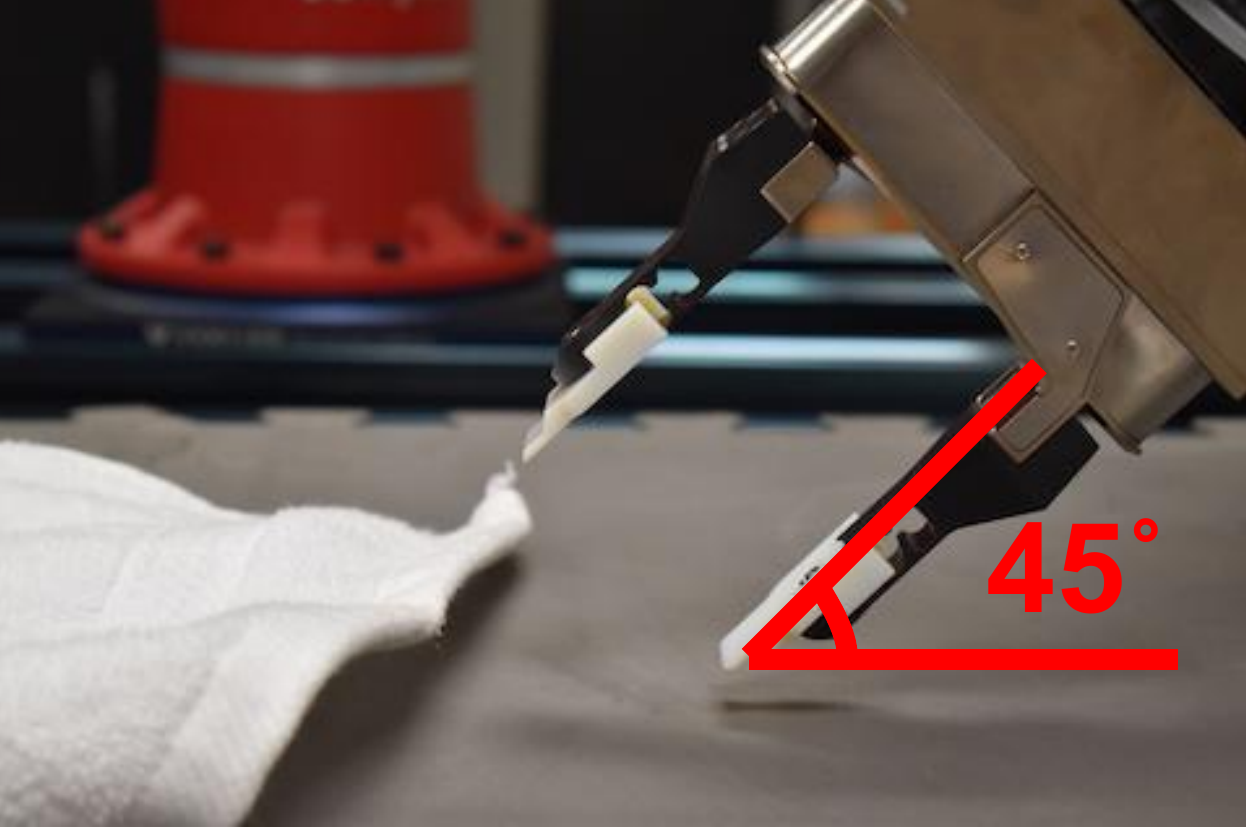}
        \caption{Pre-slide pose.}
    \end{subfigure}
    \begin{subfigure}[t]{0.32\columnwidth}
        \includegraphics[width=0.99\linewidth]{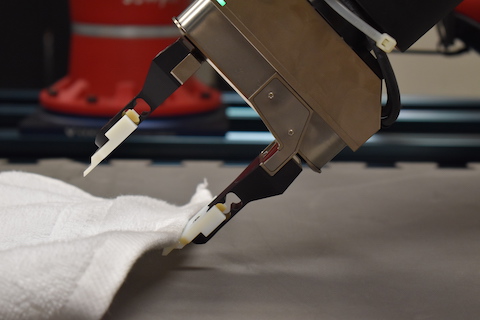}
        \caption{Post-slide pose.}
    \end{subfigure}
    \begin{subfigure}[t]{0.32\columnwidth}
        \includegraphics[width=0.99\linewidth]{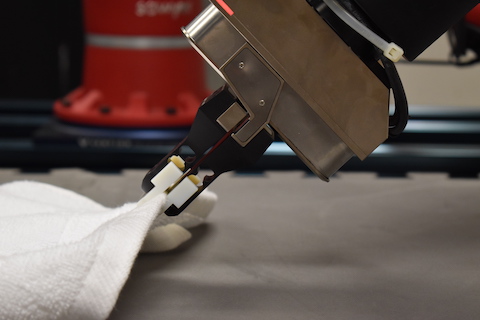}
        \caption{Pinch grasp.}
    \end{subfigure}
    \caption{Sequence of poses for the sliding grasp policy. The sliding action is a translation from the pre-slide to post-slide pose. The slide intercepts the target grasp point on the cloth.}
    \label{fig:expsetup}
\end{figure}

\subsection{Grasp Execution}
\label{sec:graspexecution}
Once a grasp configuration with point and direction is chosen, we execute a hand-designed grasping policy to slide one of the gripper's fingertips under the cloth for a pinch grasp.
We use this sliding grasp policy instead of a simpler top-down grasping routine, because top-down pinch grasps on edges and corners that are folded over (and hence overlap parts of the cloth) often result in grasping multiple layers of the cloth.
A tilted sliding grasp can separate one layer of cloth from another.

The configuration ($\mathbf{p}$, $\alpha$) specifies the grasp point on the cloth and the direction for the sliding grasp.
This configuration is specified in image coordinates; to transform it into the world frame, we perform a 2D-to-3D projection using known camera intrinsics and extrinsics. 
This provides an intermediate 6D pre-grasp pose $\mathbf{\tilde{g}}$ consisting of the 3D position of the target cloth point (corresponding to $\mathbf{p}$ in 2D), and the 3D orientation of the end-effector (corresponding to $\alpha$ in 2D). The intermediate pre-grasp pose $\mathbf{\tilde{g}}$ is oriented top-down and rotated about the $z$-axis in the world frame.
We apply a final transformation that tilts the grasp pose about the horizontal $x$-axis by 45-degrees to obtain a new pre-grasp pose $\mathbf{g}$. This pose allows one of the fingertips to get under the cloth during the slide action.
This transformation also includes a $z$-offset to account for the $z$-height of the gripper tip lowering due to the rotation.
Finally, we compute offsets to $\mathbf{g}$ in the $xy$ plane parallel to the workspace to get pre-slide and post-slide poses.
As shown in Fig.~\ref{fig:expsetup}, the sliding grasp policy moves to the pre-slide pose, translates to the post-slide pose, then pinches to grasp the cloth. 

\subsection{Implementation Details}
\label{sec:setup}

\subsubsection{Network Implementation Details}

To train our segmentation network, we collected a dataset of paired RGB-D images. The images were extracted from RGB-D video of a human manipulating a cloth with regions of interest labeled using acrylic paint. The cloth was square, 12 inches each side, and painted with red 3$\times$3 cm corners, yellow 1.5 cm thick outer edges, and green 1.5 cm thick inner edges. See Fig.~\ref{fig:training} for an image of the labeled cloth.

The human manipulated this semantically labeled cloth in the robot's workspace by folding it, dropping it, bunching it up, etc. We collected 8 minutes of video for a total of about 6700 RGB-D images. These images were split into 4:1:1 train, validation, and test sets.

Our segmentation network is based on U-Net~\cite{ronneberger2015u}.
We augmented the data during training with random image flips and rotations to improve robustness.
\hl{Additional details on training and the network architecture are provided in the appendix}.
All training was performed on an Ubuntu 16.04 machine with an NVIDIA GTX 1080 Ti GPU, a 2.1 GHz Intel Xeon CPU, and 32 GB RAM.

\subsubsection{Physical Implementation Details}
\label{sec:expsetup}

All experiments were performed on a 7 DOF Rethink Robotics Sawyer Robot with a Weiss WSG-32 parallel-jaw gripper. The robot's workspace was a 0.6$\times$0.6 m area. A Microsoft Azure Kinect sensor was mounted 0.7 m above the workspace to provide RGB-D images. Our test cloth is a white, unlabeled cloth with the same dimensions as the labeled one used for training. See Fig.~\ref{fig:overview}a for the complete workspace setup.
The default fingertips of the Weiss gripper were too thick to get under the cloth during the sliding maneuver, so we 3D-printed and attached thinner fingertips (see Fig.~\ref{fig:expsetup}).

\section{Experiments}

Our experiments are designed to answer the following questions:
\begin{itemize}
    \item How does our learned method for finding cloth edges and corners compare to non-learned baselines?
    \item How does our method for estimating the grasp direction compare to non-learned baselines?
    \item Do we obtain more robust grasps using our method for estimating grasp directional uncertainty?
\end{itemize}

\subsection{Experimental Design}

We designed two experiments to evaluate our method against various baselines. The first experiment evaluated performance for grasping cloth edges (as opposed to creases or folds), and the second evaluated grasping cloth corners. In both experiments, each grasping trial starts with a randomly crumpled cloth in the center of the robot's workspace. To enable reproducibility of our results, we used the following protocol in all of our experiments to generate the initial cloth configuration for each trial: at the beginning of each trial, a human grasps the square cloth at the midpoint of an edge. They then hold the cloth at a height such that the lowest point of the cloth is 0.1 m from the workspace surface. Finally, they let go of the cloth from this height to obtain a randomly crumpled cloth pose. This initialization procedure is based on the protocol from~\cite{garcia2020benchmarking}, adapted for our cloth grasping task.

\begin{figure}[h]
    \centering
    \begin{subfigure}[b]{0.32\columnwidth}
        \includegraphics[width=0.99\linewidth]{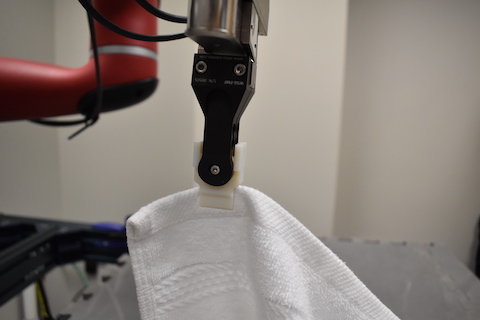}
        \caption{No fold.}
    \end{subfigure}
    \begin{subfigure}[b]{0.32\columnwidth}
        \includegraphics[width=0.99\linewidth]{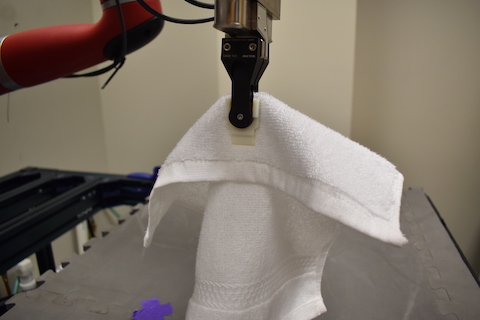}
        \caption{Single fold.}
    \end{subfigure}
    \begin{subfigure}[b]{0.32\columnwidth}
        \includegraphics[width=0.99\linewidth]{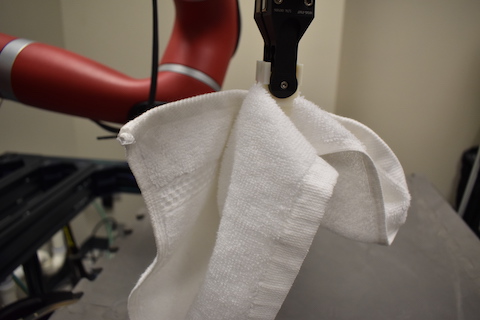}
        \caption{Multiple folds.}
    \end{subfigure}
    \caption{Examples of cloth grasps. Folds longer than 2cm from edge to fold are considered grasp failures; of these three, only (a) is considered a success.}
    \label{fig:clothfold}
\end{figure}

We define success metrics for grasping the cloth at edges and corners.
A grasp is considered a success if it pinches a cloth edge or corner and lifts it 30 cm above the workspace. 
The flexible and deformable nature of cloth can cause pinch grasps on edges and corners to fold over some of the material.
Fig.~\ref{fig:clothfold} shows examples of grasps with flat and folded cloth. 
For grasping edges, we consider a grasp with cloth folded over to be a success if the fold is less than or equal to 2 cm at its maximum length.
For grasping corners, we use a threshold of 5 cm from the corner to the fold. 
These thresholds apply when there is a single cloth fold pinched; if multiple folds are held within the pinch grasp, the grasp is considered a failure.

\subsection{Experimental Results}
\label{sec:expresults}

We evaluate whether our learned method performs better than baselines for identifying cloth edges and corners (as opposed to wrinkles and folds). Our method consists of the cloth region segmentation network, grasp direction estimation, grasp directional uncertainty estimation, and grasp selection, as described in Sec.~\ref{sec:approach}.

\subsubsection{Grasping Cloth Edges}

\begin{figure*}[ht]
    \centering
    \vspace{0.4em}
    \begin{subfigure}[t]{0.19\textwidth}
        \includegraphics[clip,trim=35px 25px 2px 5px, width=0.99\linewidth]{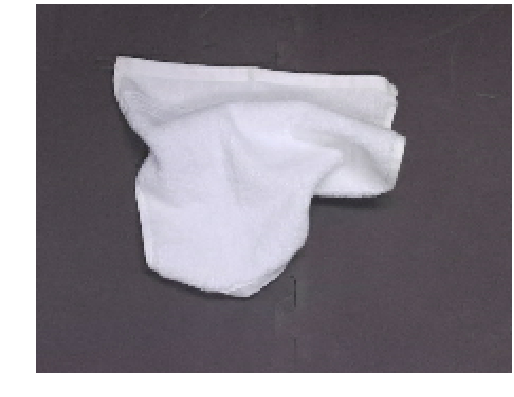}
        \caption{Cloth Pose (for reference).}
    \end{subfigure}
    \begin{subfigure}[t]{0.19\textwidth}
        \includegraphics[clip,trim=35px 25px 2px 5px, width=0.99\linewidth]{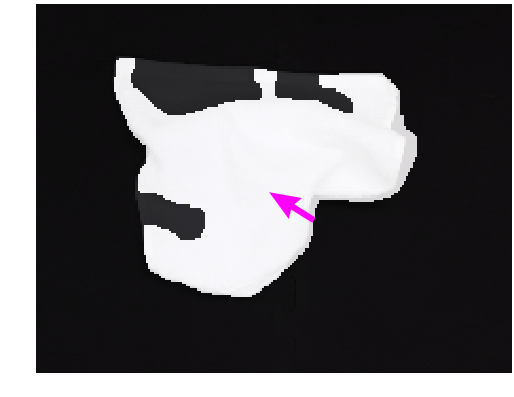}
        \caption{Segment-Edge.}
    \end{subfigure}
    \begin{subfigure}[t]{0.19\textwidth}
        \includegraphics[clip,trim=35px 25px 2px 5px, width=0.99\linewidth]{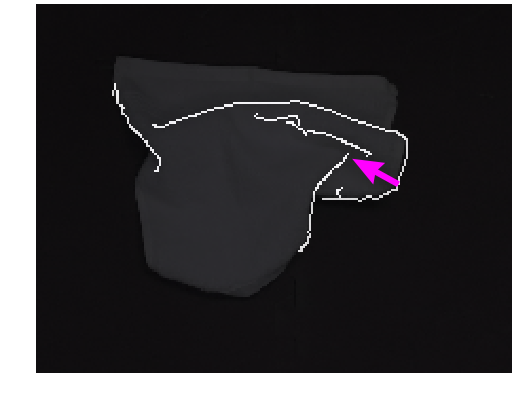}
        \caption{Canny-Depth~\cite{canny1986computational}.}
    \end{subfigure}
    \begin{subfigure}[t]{0.19\textwidth}
        \includegraphics[clip,trim=35px 25px 2px 5px, width=0.99\linewidth]{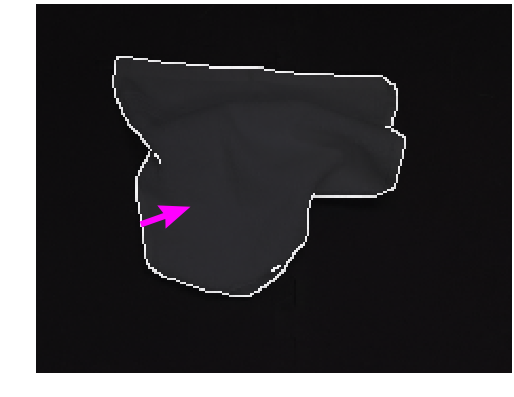}
        \caption{Canny-Color~\cite{canny1986computational}.}
    \end{subfigure}
    \begin{subfigure}[t]{0.19\textwidth}
        \includegraphics[clip,trim=35px 25px 2px 5px, width=0.99\linewidth]{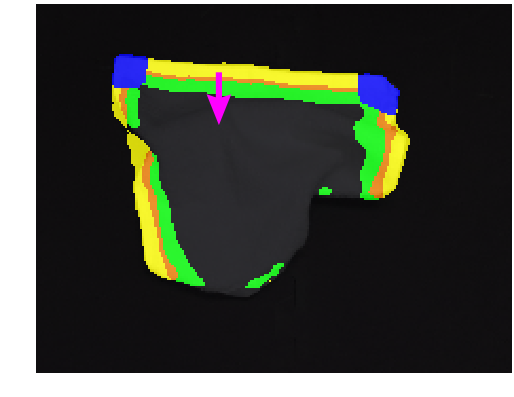}
        \caption{Our Method.}
    \end{subfigure}
    \caption{Segmentation and selected grasp point for edge grasping methods. (b)-(e) visualize the output of each method on top of the reference image (a). Note that the color image is only provided as input to Canny-Color (d); all other methods take the corresponding depth image as input. As shown in (e), our method correctly identifies most of the apparent edges of the cloth as folds, whereas the other methods fail to make this distinction.
    }
    \label{fig:methods}
\end{figure*}

For the task of identifying cloth edges, we evaluate against three baselines: 
\begin{itemize}
    \item ``Segment-Edge'' segments the cloth from the table using RANSAC plane fitting. A grasp point is randomly selected from the edge pixels of the segmentation. The grasp direction is determined by the direction of the depth gradient at the selected grasp point.
    \item ``Canny-Depth'' applies Canny edge detection~\cite{canny1986computational} to the depth image. The grasp point is sampled uniformly from the set of edge points above an intensity threshold. The grasp direction is determined by the depth gradient direction, as in the above. 
    \item ``Canny-Color'' is the same as Canny-Depth, except it applies Canny edge detection to the gray-scaled color image. The grasp direction is determined by the color gradient direction instead of depth.
\end{itemize}
See Fig.~\ref{fig:methods} for visualizations of these methods. 

The results are shown in Table~\ref{table:edges}. We performed 3 trials with 10 grasps each to estimate a mean and variance for each method. Our method significantly outperforms the baselines in terms of grasp success. The network is largely able to correctly distinguish between edges and folds, determine an appropriate grasp configuration direction, and execute a successful grasp. 
Averaging over the trials, there were an average of 2.7 failures out of 10 grasps due to misdetection, meaning that the grasp point selected was not a real edge. 
There was an average of 0.3 failures out of 10 grasps due to failed grasping. See Sec.~\ref{sec:failcases} for more details on failure cases.

The baselines perform poorly largely due to an inability to distinguish between real cloth edges versus folds. Canny-Depth relies on the intensity of depth gradients to find cloth edges, but depth gradients occur for both cloth edges and large folds. Segment-Edge fails due to noisy segmentation; because the cloth is thin, parts of the cloth can fall within the inlier threshold of the RANSAC table segmentation, despite careful parameter tuning. Still, even with a clean segmentation, grasping at an edge point on the segmentation mask often results in grasping a cloth fold for our highly crumpled cloth configurations. Canny-Color uses color gradients to find edges. It is less affected by noise compared to the depth-based baselines, as the white cloth stands out from the darker background of the table, resulting in strong edges. However, this method is still unable to discriminate between real cloth edges from folds, resulting in failure in a majority of grasp attempts. 

Our network is able to perform better than all of these baselines by using a learned segmentation. The successful grasps are also of higher quality, meaning that the grasps are more often flat with no folding of the cloth, and the edge is horizontal to the gripper tip. 
In terms of execution time, the perception component of our method runs in approximately 0.25s, with the segmentation network contributing approximately 0.14s to that total. Grasp execution is a larger bottleneck and requires approximately 15s for all methods.

\begin{table}[h]
    \centering
    \caption{Grasping Cloth Edges}
    \label{table:edges}
    \normalsize
    \begin{tabular}{l ccc}
      \toprule
        Method & Grasp Success & \\
        \midrule
        Canny-Depth & 
            $0.20\pm0.00$\\
        Segment-Edge & 
            $0.30\pm0.00$\\
        Canny-Color & 
            $0.33\pm0.12$\\
        Our Method &
            $\mathbf{0.70\pm0.20}$\\
      \bottomrule
    \end{tabular}
    \vspace{5pt}
    \caption*{3 trials per method, 10 grasp attempts per trial}
\end{table}

\subsubsection{Grasping Cloth Corners}

We also evaluated our method on grasping corners. Our method remains the same, except that corners are used for grasp point selection instead of edges. The corners still use correspondence with inner edges to determine grasp direction, as well as our method for estimating grasp directional uncertainty described in Sec.~\ref{sec:approach}.

For this task, we evaluated against the following baselines:
\begin{itemize}
    \item ``Harris-Depth'' applies Harris corner detection~\cite{harris1988combined} to the depth image. The maximum intensity value is selected as the grasp point. The depth gradient direction at the grasp point is used to determine the grasping direction, as in the edge grasping experiments. 
    \item ``Harris-Color'' takes a grayscaled RGB image as input and uses color gradients to determine the grasping direction, but is otherwise the same as the above. 
\end{itemize} 

The results are shown in Table~\ref{table:corners}. Our method outperforms the baselines on corner grasping, being able to more reliably detect corners in any cloth configuration.
Averaging over the trials, there were an average of 3 failures out of 10 grasps due to misdetection.
There were an average of 1.3 failures out of 10 grasps due to grasping error. Our method performs worse on corners than on edges. Fewer regions of the image are corners compared to edges, so false positives are more problematic. Sec.~\ref{sec:failcases} for details on failure cases.

The baselines perform poorly for largely the same reason of misdetection as with the edge experiments. The Harris-Depth baseline performs poorly because it looks for large changes in the gradient in all directions, which could result in false positives instead of real corners. Most of the grasp point selections from this baseline were on wrinkles and folds than on the cloth. The Harris-Color baseline performs better than depth, possibly because there are fewer false positives given the white on black input images. White cloth corners against the darker workspace surface can be easily detected; however, corners lying on top of the cloth are less likely to be detected. For our difficult randomly crumpled cloth configurations, the corners are not always cleanly visible against the surface, and often lie in configurations that are difficult to discriminate in 2D. 

\begin{table}[h]
    \centering
    \caption{Grasping Cloth Corners}
    \label{table:corners}
    \normalsize
    \begin{tabular}{l ccc}
      \toprule
        Method & Grasp Success \\
        \midrule
        Harris-Depth & 
            $0.05\pm0.07$\\
        Harris-Color & 
            $0.33\pm0.15$\\
        Our Method &
            $\mathbf{0.57\pm0.06}$\\
      \bottomrule
    \end{tabular}
    \vspace{5pt}
    \caption*{3 trials per method, 10 grasp attempts per trial}
    \vspace{-1em}
\end{table}

\subsubsection{Ablations}

We perform ablations on our method to determine the relative contribution of the different components of our method to grasp success. Our full method consists of segmenting cloth regions using a neural network (Sec.~\ref{sec:clothseg}), determining the grasp direction for all segmented edge/corner pixels using their nearest segmented inner edge pixels (Sec.~\ref{sec:graspdir}), and selecting a grasp point with the lowest grasp directional uncertainty (Sec.~\ref{sec:diruncertainty}).

We perform the following ablations of our method:
\begin{itemize}
    \item ``No-Direction-Prediction'' still uses the cloth segmentation network of Sec.~\ref{sec:clothseg}. However, rather than determining the grasp direction using the methods of Sec.~\ref{sec:graspdir} and Sec.~\ref{sec:diruncertainty}, this ablation determines the grasp direction by fitting a bounding box around the segmented outer edge pixels and setting the direction to be the vector pointing to the center of the box. Instead of using the point with minimum directional uncertainty, it randomly selects the grasp point from the set of outer edge pixels. 
    \item ``No-Directional-Uncertainty'' still uses the cloth segmentation network of Sec.~\ref{sec:clothseg} as well as the method of Sec.~\ref{sec:graspdir} for determining the grasp direction.  However, rather than computing the grasp directional uncertainty to choose a grasp point as in Sec.~\ref{sec:diruncertainty}, this ablation chooses a grasp point randomly.
\end{itemize} 

The results are shown in Table~\ref{table:ablations}.
The ablations under-perform the full method,  demonstrating that our method for estimating the grasp direction (Sec.~\ref{sec:graspdir}) as well as our method for estimating directional uncertainty (Sec.~\ref{sec:diruncertainty}) help to choose more robust grasps. 
We observe No-Direction-Prediction selecting grasp directions near-parallel to real edges instead of orthogonally, because it always chooses directions toward the center of the segmentation bounding box.
The performance of No-Directional-Uncertainty vs.\ No-Direction-Prediction provides evidence that using the inner edge segmentation to determine the grasp direction improves grasp success.
Comparing our full method with No-Directional-Uncertainty shows that selecting the grasp point with minimal directional uncertainty outperforms random grasp point selection. 

\begin{table}[h]
    \centering
    \caption{Ablations on Grasping Cloth Edges}
    \label{table:ablations}
    \normalsize
    \begin{tabular}{l ccc}
      \toprule
        Method & Grasp Success \\
        \midrule
        No-Direction-Prediction & 
            $0.2$\\
        No-Directional-Uncertainty & 
            $0.4$\\
        Our Method &
            $\mathbf{0.7\pm0.20}$\\
      \bottomrule
    \end{tabular}
    \vspace{5pt}
    \caption*{1 trial per ablation, 10 grasp attempts in trial}
\end{table}

\subsubsection{Failure Cases}
\label{sec:failcases}

\begin{figure}[ht]
    \centering
    \begin{subfigure}[t]{0.32\columnwidth}
        \includegraphics[clip,trim=35px 60px 2px 5px, width=0.99\linewidth]{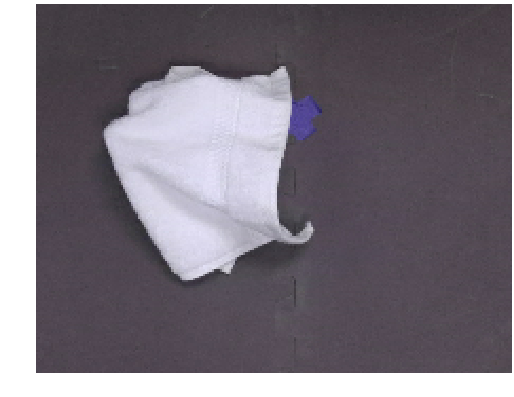}
    \end{subfigure}
    \begin{subfigure}[t]{0.32\columnwidth}
        \includegraphics[clip,trim=35px 60px 2px 5px,width=0.99\linewidth]{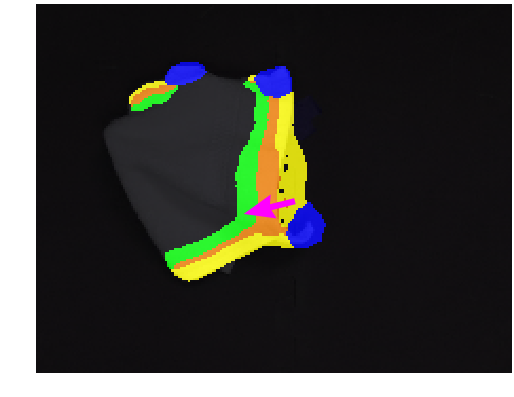}
    \end{subfigure}
    \begin{subfigure}[t]{0.32\columnwidth}
        \includegraphics[width=0.99\linewidth]{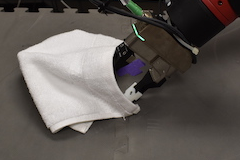}
    \end{subfigure}\\
    \vspace{2pt}
    
    \begin{subfigure}[t]{0.32\columnwidth}
        \includegraphics[clip,trim=35px 60px 2px 5px,width=0.99\linewidth]{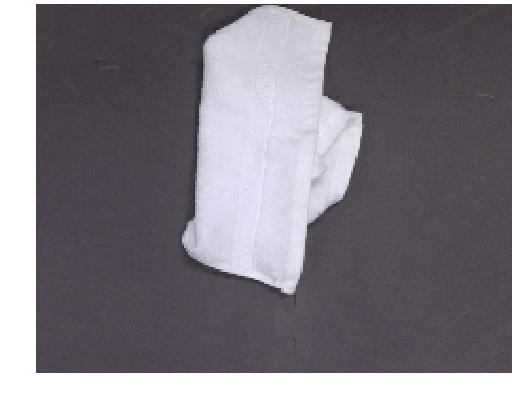}
        \caption{RGB Image.}
    \end{subfigure}
    \begin{subfigure}[t]{0.32\columnwidth}
        \includegraphics[clip,trim=35px 60px 2px 5px,width=0.99\linewidth]{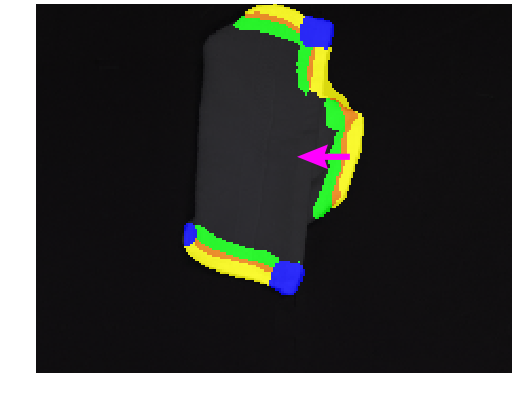}
        \caption{Segmentation and Grasp Prediction.}
    \end{subfigure}
    \begin{subfigure}[t]{0.32\columnwidth}
        \includegraphics[width=0.99\linewidth]{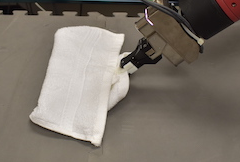}
        \caption{Grasp execution.}
    \end{subfigure}
    \caption{Failure cases. (top row) Segmentation bleeds over real cloth edge, leading to poor estimation of grasp height. (bottom row) Grasp fails to avoid grasping nearby folds and edges (note that misdetection has also occurred).}
    \label{fig:fails}
\end{figure}

In this section we discuss the most frequent and notable failure cases. Examples of these cases are in Fig.~\ref{fig:fails} and the supplementary video. 

Failures occurred when the segmentation produced by our method contained errors. Because the cloth is very thin and the depth images captured from our sensor are noisy, the network can fail to get accurate segmentation at cloth edges (see Fig.~\ref{fig:fails}, top row).
This issue causes both false positives, in which pixels close to real edges are included in the segmentation, and false negatives, in which the segmentation does not include valid pixels.
These segmentation errors affect the grasp selection component that takes the segmentation as input. 
As a result, we sometimes observed our method selecting grasp points on false positives, which were more likely to result in grasp failures.

Failures also occurred due to grasping areas with valid edges but problematic nearby cloth configurations.
For example, overlapping edges can create the appearance of a continuous segmentation, and a grasp on that area will result in grasping both edges (see Fig.~\ref{fig:fails}, bottom row). 
Developing a policy that can adapt to such challenging configurations is an area of future work.

Failures due to motion planning to reach commanded poses happened infrequently, such as when a selected grasp is in an unreachable robot configuration. These failures are easily detected, so we re-execute our method to choose a different grasp point in such cases.

\subsubsection{Robustness}
\label{sec:robust}

\begin{figure}[ht]
    \centering
    \begin{subfigure}[t]{0.49\columnwidth}
        \includegraphics[
        clip,trim=0px 60px 0px 0px, 
        width=0.99\linewidth]{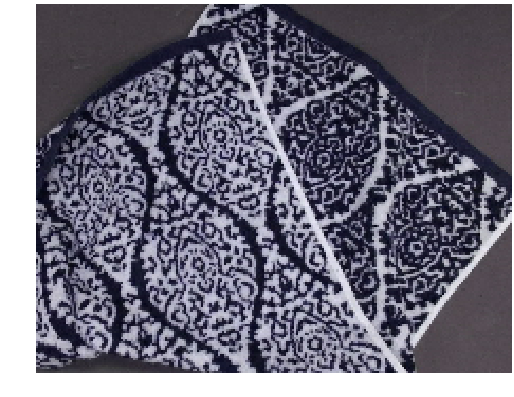}
    \end{subfigure}
    \begin{subfigure}[t]{0.49\columnwidth}
        \includegraphics[
        clip,trim=0px 60px 0px 0px,
        width=0.99\linewidth]{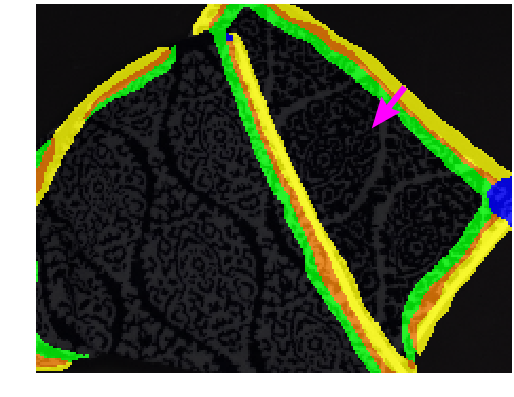}
    \end{subfigure}
    \vspace{0.05em}
    
    \begin{subfigure}[t]{0.49\columnwidth}
        \includegraphics[
        clip,trim=0px 60px 0px 0px,
        width=0.99\linewidth]{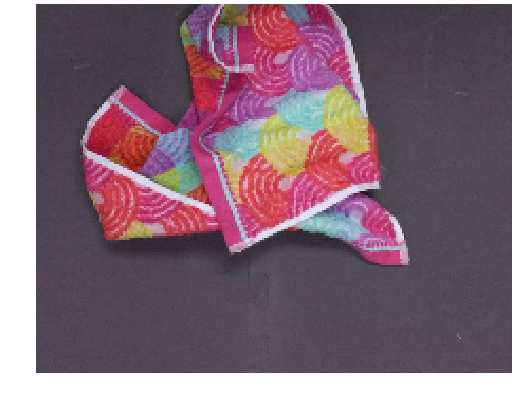}
        \caption{RGB Image.}
    \end{subfigure}
    \begin{subfigure}[t]{0.49\columnwidth}
        \includegraphics[
        clip,trim=0px 60px 0px 0px,
        width=0.99\linewidth]{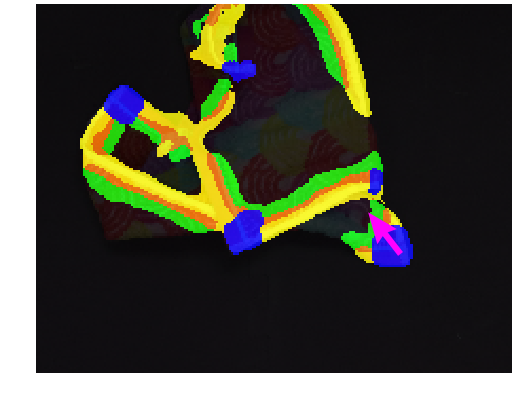}
        \caption{Segmentation and Selected Grasp.}
    \end{subfigure}
    \caption{Our network is able to segment cloths of various sizes and visual texture. See the supplementary video for grasping demonstrations on these cloths.}
    \label{fig:robust}
    \vspace{-1em}
\end{figure}

\hl{We demonstrate that our network is robust to variations in visual texture and cloth size by grasping other cloths (see Fig.}~\ref{fig:robust}\hl{ and supplementary video). Our network can segment cloth with different colors and patterns because it only takes depth as input. It can also segment cloths of different dimensions due to its fully convolutional architecture.}

\section{Conclusion} 
\label{sec:conclusion}

We present a method to segment real edges and corners of cloth (as opposed to creases or folds) from depth images. 
Our method also determines a grasp configuration from these segmentations that accounts for directional uncertainty. 
We demonstrate a system that implements our approach to grasp cloths in crumpled configurations, and we show that our method outperforms various baselines in terms of grasp success rate on grasping success.






\section*{ACKNOWLEDGMENT}

This work was supported by the National Science Foundation Smart and Autonomous Systems Program (IIS-1849154), the United States Air Force and DARPA under Contract No. FA8750-18-C-0092, LG Electronics, a NSF Graduate Research Fellowship (DGE-1745016),  and a NASA Space Technology Research Fellowship (80NSSC17K0233).


\bibliographystyle{IEEEtran.bst}
\bibliography{root}

\clearpage
\begin{appendix}

\subsection{Network Architecture}

Our network architecture is based on U-Net~\cite{ronneberger2015u}. It consists of a downsampling part and an upsampling part. In the downsampling path, a step consists of two 3x3 unpadded convolutions, each with batch normalization and a rectified linear unit, followed by a 2x2 max pooling layer with stride 2. We apply four of these steps, doubling the number of feature channels each time. For the upsampling path, a step consists of a 2x2 up-convolution that halves the number of feature channels, a concatenation with a cropped feature map from the corresponding downsampled path, and two 3x3 convolutions, each followed by batch normalization and ReLU. A final 1x1 convolution is used to turn the feature map into 3 classes for corners, outer edges, and inner edges respectively. 

The differences between our network and U-Net are that we add batch normalization, and our network takes a single channel depth image as input. 

\subsection{Network Training}
\label{appendix:hyperparams}

We implemented the network in PyTorch. We use the Adam optimizer with a learning rate of 1e-5. We use a batch size of 8. To augment our data, we flip the image with 50 percent chance, and also rotate the image with 50 percent chance, sampling within [-30 degrees, 30 degrees]. 

In our loss function, we set the per-class (corners, outer edges, and corners) weight $w_k$ for balancing the loss on positive and negative predictions to 20 for all classes. 

\subsection{Grasp Direction Uncertainty Estimation}

As described in Sec.~\ref{sec:graspconfig}.2, the uncertainty of the grasp direction for a single outer edge point $\mathbf{p}$ is the variance of the grasp directions predicted by its neighbors. Each neighbor is an outer edge pixel with its own grasp direction vector, computed as described in Sec.~\ref{sec:graspconfig}.1. We form the neighborhood by taking the $k$ outer edge pixel points closest to $\mathbf{p}$, and set $k=100$.  

\end{appendix}

\end{document}